\documentclass[conference]{IEEEtran}
\IEEEoverridecommandlockouts
\usepackage{cite}
\usepackage{caption,subcaption}
\usepackage{url}
\usepackage{amsmath,amssymb,amsfonts}
\usepackage{algorithmic}
\usepackage{graphicx}
\usepackage{textcomp}
\usepackage{xcolor}
\usepackage{float}
\def\BibTeX{{\rm B\kern-.05em{\sc i\kern-.025em b}\kern-.08em
    T\kern-.1667em\lower.7ex\hbox{E}\kern-.125emX}}

    \makeatletter
\newcommand{\linebreakand}{%
  \end{@IEEEauthorhalign}
  \hfill\mbox{}\par
  \mbox{}\hfill\begin{@IEEEauthorhalign}
}
\makeatother
\begin{document}

\title{Combining Datasets with Different Label Sets for Improved Nucleus Segmentation and Classification}

\author{\IEEEauthorblockN{Amruta Parulekar*}
\IEEEauthorblockA{\textit{Dept. of Electrical Engineering} \\
\textit{Indian Institute of Technology, Bombay}\\
Mumbai, India \\
20d070009@iitb.ac.in}
\and
\IEEEauthorblockN{Utkarsh Kanwat*}
\IEEEauthorblockA{\textit{Dept. of Electrical Engineering} \\
\textit{Indian Institute of Technology, Bombay}\\
Mumbai, India \\
190070069@iitb.ac.in}
\and
\IEEEauthorblockN{Ravi Kant Gupta}
\IEEEauthorblockA{\textit{Dept. of Electrical Engineering} \\
\textit{Indian Institute of Technology, Bombay}\\
Mumbai, India \\
184070025@iitb.ac.in}
\and
\IEEEauthorblockN{Medha Chippa}
\IEEEauthorblockA{\textit{Dept. of Electrical Engineering} \\
\textit{Indian Institute of Technology, Bombay}\\
Mumbai, India \\
medha6271@gmail.com}
\and
\IEEEauthorblockN{Thomas Jacob}
\IEEEauthorblockA{\textit{Dept. of Electrical Engineering} \\
\textit{Indian Institute of Technology, Bombay}\\
Mumbai, India \\
190070068@iitb.ac.in}
\and
\IEEEauthorblockN{Tripti Bameta}
\IEEEauthorblockA{\textit{Dept. of Medical Oncology} \\
\textit{Tata Memorial Centre-ACTREC (HBNI)}\\
Mumbai, India \\
tripti.bameta@gmail.com}
\linebreakand

\IEEEauthorblockN{Swapnil Rane}
\IEEEauthorblockA{\textit{Dept. of Pathology} \\
\textit{Tata Memorial Centre-ACTREC (HBNI)}\\
Mumbai, India \\
raneswapnil82@gmail.com}
\and
\IEEEauthorblockN{Amit Sethi}
\IEEEauthorblockA{\textit{Dept. of Electrical Engineering} \\
\textit{Indian Institute of Technology, Bombay}\\
Mumbai, India \\
asethi@iitb.ac.in}
}

\maketitle
\def\thefootnote{*}\footnotetext{These authors contributed equally to this work}
\begin{abstract}
Segmentation and classification of cell nuclei in histopathology images using deep neural networks (DNNs) can save pathologists' time for diagnosing various diseases, including cancers, by automating cell counting and morphometric assessments. It is now well-known that the accuracy of DNNs increases with the sizes of annotated datasets available for training. Although multiple datasets of histopathology images with nuclear annotations and class labels have been made publicly available, the set of class labels differ across these datasets. We propose a method to train DNNs for instance segmentation and classification on multiple datasets where the set of classes across the datasets are related but not the same. Specifically, our method is designed to utilize a coarse-to-fine class hierarchy, where the set of classes labeled and annotated in a dataset can be at any level of the hierarchy, as long as the classes are mutually exclusive. Within a dataset, the set of classes need not even be at the same level of the class hierarchy tree. Our results demonstrate that segmentation and classification metrics for the class set used by the test split of a dataset can improve by pre-training on another dataset that may even have a different set of classes due to the expansion of the training set enabled by our method. Furthermore, generalization to previously unseen datasets also improves by combining multiple other datasets with different sets of classes for training. The improvement is both qualitative and quantitative. The proposed method can be adapted for various loss functions, DNN architectures, and application domains.

\end{abstract}


\section{Introduction}


Histopathology is practice of preparation of tissue slides and their examination to identify visual signs and grades of various diseases, including cancers. A surgical or biopsied tissue sample is fixed, embedded, sliced, mounted on a glass slide, and stained most commonly with hematoxylin and eosin (H\&E) to highlight various tissue components. A slide thus prepared is either observed using a high powered microscope or scanned as a gigapixel whole slide image (WSI). Tissue abnormalities can be identified using visual features, such as nucleus to cytoplasm ratio, nuclear pleomorphism, and counts of various types of cells. Usually histopathological examination relies on nuclear details for estimating these features as the cell (cytoplasmic) boundaries are not easy to identify in H\&E stained samples. Automating instance segmentation and classification of nuclei using  deep neural networks (DNNs), such as HoVerNet~\cite{graham2019hover} and StarDist~\cite{Schmidt_2018}, can bring efficiencies and objectivity to several types of histological diagnostic and prognostic tasks.
\begin{table*}[t]
\centering
\caption{Characteristics of notable nucleus segmentation and classification datasets}
\label{tab:Datasets}

 \begin{tabular}{|l|c|c|c|r|c|c|}

\hline

Dataset         & Classes                                                                               & Organs                                                                  & Mag. & Nuclei                                                    & Images                                             & Img. Size                                                      \\ 
\hline

PanNuke~\cite{gamper2020pannuke} & \begin{tabular}[c]{@{}c@{}}5:\\ Inflammatory, \\Neoplastic, \\ Dead, Connective, \\Non-neoplastic Epithelial\end{tabular}                    & \begin{tabular}[c]{@{}c@{}}19:\\ Bladder, Ovary, Pancreas, Thyroid, \\Liver, Testis, Prostrate, Stomach, \\Kidney, Adrenal gland, Skin, Head \& Neck, \\Cervix, Lung, Uterus, Esophagus, \\Bile-duct, Colon, Breast\end{tabular}             & \begin{tabular}[c]{@{}c@{}}40x\end{tabular}  & \begin{tabular}[c]{@{}r@{}}216,345\end{tabular} & \begin{tabular}[c]{@{}c@{}}481\end{tabular} & \begin{tabular}[c]{@{}c@{}}224x224\\ \end{tabular} \\ 
\hline

MoNuSAC~\cite{9446924} & \begin{tabular}[c]{@{}c@{}}4:\\ Epithelial, lymphocytes,\\ macrophages,neutrophils \end{tabular}                    & \begin{tabular}[c]{@{}c@{}}4:\\ Breast, Kidney, Liver, prostrate\end{tabular}             & 40x  & \begin{tabular}[c]{@{}r@{}}46,909\end{tabular} & \begin{tabular}[c]{@{}c@{}}310\end{tabular} & \begin{tabular}[c]{@{}c@{}}82x35\\ to\\ 1422x2162\end{tabular} \\ 
\hline


CoNSeP~\cite{graham2021conic} & \begin{tabular}[c]{@{}c@{}}7:\\ Healthy Epithelial, Inflammatory,\\ Muscle, Fibroblast,\\Dysplastic/Malignant Epithelial,\\Endothelial,Other\end{tabular}                    & \begin{tabular}[c]{@{}c@{}}1:\\ Colon\end{tabular}             & 40x  & \begin{tabular}[c]{@{}r@{}}24,319\end{tabular} & \begin{tabular}[c]{@{}c@{}}41\end{tabular} & \begin{tabular}[c]{@{}c@{}}1000x1000\end{tabular} \\ 
\hline




TNBC~\cite{naylor2018segmentation} & \begin{tabular}[c]{@{}c@{}}4:\\Basal-like, Mesenchymal, \\ Endothelial, \\ luminal androgen receptor, \\ immune-enriched\end{tabular}                    & \begin{tabular}[c]{@{}c@{}}1:\\ Breast\end{tabular}             & 40x  & \begin{tabular}[c]{@{}r@{}}4,022\end{tabular} & \begin{tabular}[c]{@{}c@{}}50\end{tabular} & \begin{tabular}[c]{@{}c@{}}512x512\end{tabular} \\ 
\hline



\end{tabular}
\end{table*}

Challenges to accurate segmentation and classification of nuclei using DNNs include intra-class variation, inter-class feature (e.g. size) overlap, the presence of physically overlapping nuclei in certain disease conditions, and the need for domain generalization. Domain differences are present due to the diversity of nuclear shapes and sizes across organs and diseases, as well as the variance in slide staining protocols and reagents and digital scanner or camera characteristics across pathology labs. Because DNNs can be scaled to generalize better with more diverse and larger datasets, it is necessary to accurately annotate and label multiple large datasets for their training. In the last few years, several annotated histological datasets have been released that differ in the sets of nuclear class labels, magnification, source hospitals, scanning equipment, organs, and diseases. For instance, while the PanNuke dataset covers 19 organs with semi-automated annotation of five nuclear classes -- neoplastic, non-neoplastic epithelial, inflammatory, connective, dead~\cite{gamper2020pannuke}; the MoNuSAC covers four organs with the following four nuclear classes -- epithelial, lymphocytes, macrophages, and neutrophils~\cite{9446924}. While most of this input diversity is beneficial to train generalized DNNs, combined training across datasets with different sets of class labels remains a challenge. Existing methods to train DNNs on multiple datasets that differ class label sets are not satisfactory. For instance, transfer~\cite{yosinski2014transferable} and multi-task learning~\cite{zhang2014facial}, do not train the last (few) layer(s) of a DNN on more than one dataset.

We propose a method to train DNNs for instance segmentation and classification over multiple related datasets for the same types of objects that have different class label sets. Specifically, we make the following contributions. (1) We propose a method to modify a wide variety of loss functions used for segmentation and classification. (2) The method is applicable whenever the class label sets across the datasets can be expressed as a part of a common coarse-to-fine class hierarchy tree. That is, the method can jointly utilize multiple datasets of the same types of objects wherein some datasets may have labels for finer sub-classes while others may have labels for coarser super-classes, or a mix of these, from the same class hierarchy tree. Apart from this type of relation among datasets, the method has no other constrains. That is, it can be used to train a wide variety of DNNs for instance segmentation and classification for various types of objects of interest, although we used the segmentation of nuclei in histopathology using StarDist~\cite{Schmidt_2018} as a case study. (3) We 
demonstrate quantitative and qualitative improvements in nuclear segmentation and classification test accuracy using the proposed method to train on multiple datasets with different class label sets. (4) We also show that thus using multiple datasets also improves domain generalization on a previously unseen dataset.



\begin{figure*}[t]
\centering
\includegraphics[clip,width=17cm,trim=0cm 9.5cm 3cm 0cm]{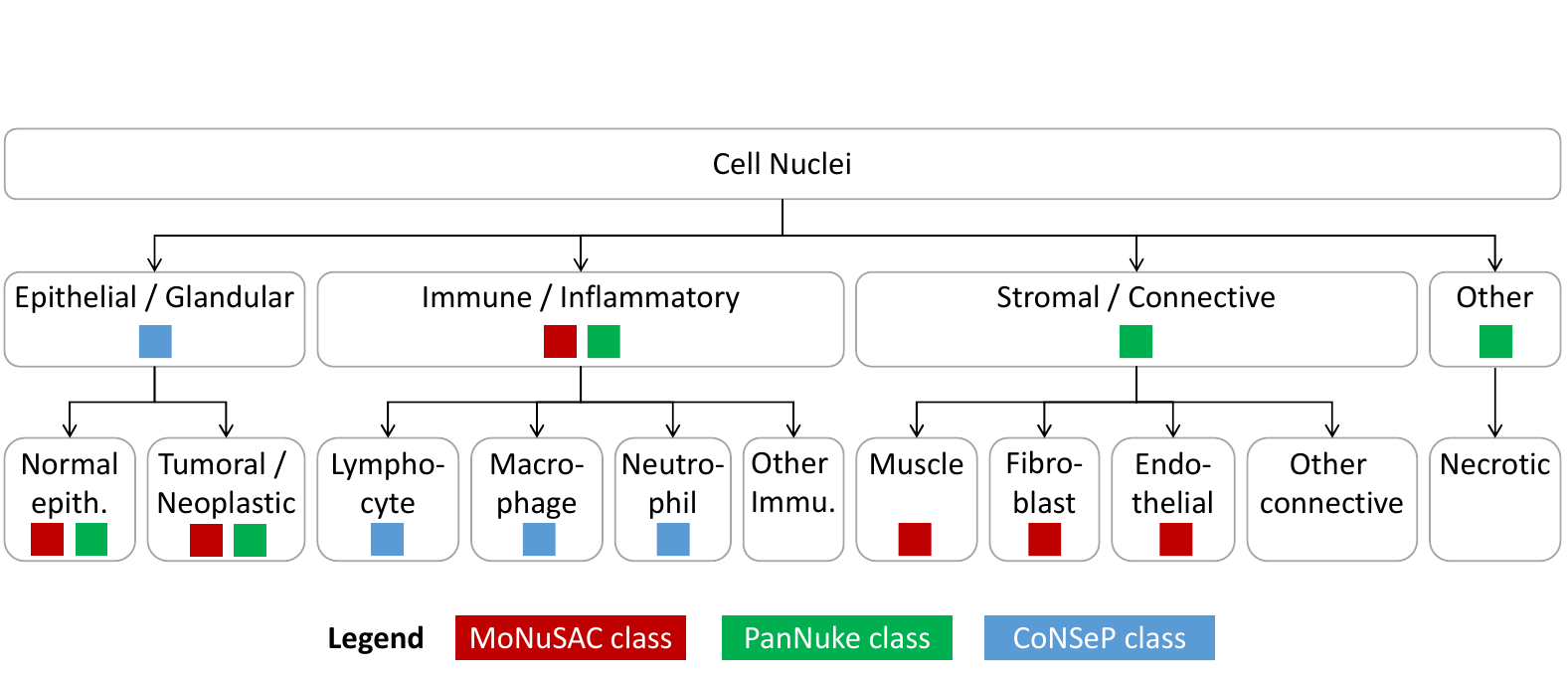}
\caption{The hierarchy of cell nucleus classes and their correspondence to the three label sets of the datasets used in this study}
\label{fig:ClassTree}
\end{figure*}

\section{Datasets, Background, and Related Work}

In this section, we review nucleus segmentation datasets and methods, and previous attempts to combine knowledge from multiple datasets.

\subsection{Nucleus segmentation datasets}
\label{sec:datasets}


Over the last few years, several datasets with careful annotations and labeling of cell nuclei have been released to the public to enable research on better instance segmentation and classification models. Some notable datasets are shown in Table~\ref{tab:Datasets}. These datasets meet our goals as they contain images with more nuclear details at 40x magnification and contain labels for nuclei from multiple classes, unlike, for example, MoNuSeg~\cite{kumar2017dataset} or CryoNuSeg~\cite{mahbod2021cryonuseg}.

\subsection{Nucleus instance segmentation and classification methods}


Over the years several DNN architectures have been developed to segment nuclei. These either use state-of-the-art image classification DNNs, such as ResNets\cite{he2016deep}, VGGnets~\cite{simonyan2014very}, and EfficientNet\cite{tan2019efficientnet} as backbones for feature extraction or finetune derivatives of U-Net~\cite{ronneberger2015u}. 
But owing to their poor generalizability and adaptability over a task as specialized as nucleus segmentation, their usage as backbone architectures has recently decreased. These have been replaced by the development of combination architectures (fusion of multiple networks) and specialized architectures. For instance, HoVerNet~\cite{graham2019hover} was proposed to predict whether a pixel location is inside a nucleus and its horizontal and vertical distances from the nuclear boundary. This concept has been generalized to predict multi-directional distance using StarDist~\cite{Schmidt_2018}. These architectures are specifically designed for histological images with overcrowded nuclei and have demonstrated state-of-the-art results compared to previous methods, such as mask R-CNN~\cite{he2018mask} or nucleus boundary mapping~\cite{kumar2017dataset}. 

\subsection{Previous attempts to use multiple datasets}

In order to combine knowledge from multiple datasets, transfer and multi-task learning have been proposed for natural and medical images. 
For instance, \cite{reis2023transfer} proposes a transfer learning technique using the MedCLNet database. DNNs were pre-trained through the proposed method and were used to perform classification on the colorectal histology MNIST dataset. The GSN-HVNET\cite{zhao2023gsn} was proposed with an encoder-decoder structure for simultaneous segmentation and classification of nuclei, and was pre-trained on the PanNuke dataset.


Although coarse-to-fine class structure has been exploited for knowledge transfer in other domains~\cite{li2019exploiting}, it has not been used in medical datasets for increasing the available data for training or for domain generalization. 
All the methods described so far have only dealt with the scenario of carrying out segmentation and classification by splitting the same dataset into training and testing, or using the same set of classes across training and testing. At best, transfer learning was carried out where only the lower pretrained layers were retained and new upper layers were randomly initialized and trained on target datasets. 
There are no loss functions or training methods that can train the entire DNN on multiple datasets to utilize them for all layers as well as for cross-domain (dataset) generalization of segmentation and classification.


\section{Proposed Method}

We propose a method to train DNNs for segmentation and classification on multiple datasets with related but potentially different class label sets. We assume that the class label sets across the datasets are different cuts of the same class hierarchy tree. Within each dataset, the class labels are mutually exclusive, but need not be collectively exhaustive. An example of a class hierarchy tree with different cuts for labels for three different datasets is given in Figure~\ref{fig:ClassTree}, where nuclei can be divided into four super-classes, which in turn can be divided into 11 sub-classes. Deeper and wider hierarchies can also be used. Class label sets that are not a part of a common class hierarchy tree are out of the scope of this work.


Our key idea is to modify a class of loss functions whose computation involves sums over predicted and ground truth class probability terms in conjunction with sums over instances or pixels. This description covers a wide array of loss functions, including cross entropy, Dice loss, focal loss, Tversky loss, focal Tversky loss~\cite{abraham2018novel}. As an astute reader might have guessed by now, we propose to sum the predicted probabilities of fine-grained sub-classes when the class label can only be given for their coarser super-class. The set of finer sub-classes to be combined using such a method of loss computation can even dynamically change from dataset-to-dataset, epoch-to-epoch, batch-to-batch, or even instance-to-instance. To keep things simple, we first train the model on one dataset for a few epochs, and then train it on a second dataset for the remaining epochs.

This method is also applicable to any DNN architecture or application domain (e.g., natural images) that can be trained using these losses. As a case study, we use it to modify cross entropy and focal Tversky loss functions~\cite{abdolhoseini2019segmentation} to train a UNet-based StarDist DNN~\cite{Schmidt_2018} for H\&E-stained histopathology nucleus segmentation and classification on MoNuSAC, PanNuke, and CoNSeP datasets.

Although this method can be extended to multiple levels, for simplicity of explanation we will assume that a class label can be at one of the two levels -- a super-class or a sub-class. We design a neural architecture that makes predictions at the finest level of the hierarchy, which is the set of all sub-classes (plus background) in this case. 
When the label for a training instance is available at the super-class level, we add the predicted probabilities of its sub-classes, and update their weights with an equal gradient, as should be done backwards of a sum node. This way, the weights leading to the prediction of all sub-classes are trained even when only the super-class label is available. The gradient and output obtained from this approach is at the finest (sub-class) level, but we interpret the results for a dataset only for its corresponding training label set. That is, we do not assess sub-class level performance when only super-class labels are available, even though we train the DNN to predict at the sub-class level. 
On the other hand, when we come across a training instance where a sub-class label is available, we skip the sum-based merging of probability masses. In this case, class-specific weight update and the interpretation of predictions proceeds in the usual fashion. 




Consider the cross entropy loss for a fixed set of class labels:
\begin{equation}
L_{CE}=-\sum_{i=1}^{n} \sum_{j=1}^{c} t_{ij}  \log(y_{ij}),
\label{eq:CE}
\end{equation}
where $n$ is the number of training instances, $c$ is the number of classes, $t_{ij}$ are one-hot labels, and $y_{ij}$ are the predicted class probabilities such that $\forall i \sum_{j}t_{ij}=1, \sum_{j}y_{ij}=1$. In case a subset of classes belong to a super-class $k$ denoted by $j \in S_k$, then we modify Equation~\ref{eq:CE} as follows:
\begin{equation}
L_{MCE}=-\sum_{i=1}^{n} \sum_{k=1}^{m} t_{ik} \log\left(\sum_{j \in S_k} y_{ij}\right),
\label{eq:SumCE}
\end{equation}
where $t_{ik}$ is a binary indicator label for the super-class $k$, and $m$ is the size of the class label set. That is, $t_{ik} = \sum_{j \in S_k} t_{ij}$, but the individual terms $t_{ij}$ may not be known in the given labels. As is clear from Equation~\ref{eq:SumCE}, that although for notational simplicity, the sum over classes runs at the super-classes enumerated by $k$ at the same level, the modification applies independently to each branch of the class hierarchy tree (see Figure~\ref{fig:ClassTree} for an example), as was done in our implementation. Additionally, it is also clear that the method can be extended to deeper and wider hierarchy trees with label sets that are arbitrary cuts of the tree.

We next consider a slightly more complex loss -- the focal Tversky loss~\cite{abraham2018novel}:
\begin{equation}
\begin{aligned}
L_{FT}=\sum_{i=1}^{n} \left(1 - \frac{{\sum_{j=1}^{c}{(t_{ij}}  y_{ij}} + \epsilon)}{\alpha  \sum_{j=1}^{c} t_{ij} + (1 - \alpha)  \sum_{j=1}^{c} y_{ij} + \epsilon} \right)^\gamma,        
\end{aligned}
\label{eq:FT}
\end{equation}
where $\epsilon$ is a small constant to prevent division by $0$, and $\alpha>0,\gamma>0$ are hyper-parameters. Following the same principles as used to propose the loss in Equation~\ref{eq:SumCE}, we now propose a modified focal Tversky loss:
\begin{equation}
\begin{split}
\begin{aligned}
L_{MFT}= \hspace{7cm} \\
\sum_{i=1}^{n} \left(1 - \frac{{\sum_{k=1}^{m}{(t_{ik}} \sum_{j \in S_k}
y_{ij}} + \epsilon)}{\alpha \sum_{k=1}^{m} t_{ik} + (1 - \alpha) \sum_{k=1}^{m} \sum_{j \in S_k} y_{ij} + \epsilon}\right)^\gamma.
\end{aligned}
\end{split}
\label{eq:SumFT}
\end{equation}
Once again, it is clear that $L_{MFT}$ can also be modified to be applied independently to each branch and sub-branch of a class hierarchy tree, including labels at different levels of the tree that are in different branches.

In our implementation of nuclear instance segmentation and classification, we used a positive combination of $L_{MCE}$ and $L_{MFT}$ with suitable modifications of Equations~\ref{eq:SumCE} and~\ref{eq:SumFT} to handle classes at different levels, as shown in Figure~\ref{fig:ClassTree}. These are modified versions of the losses used in the orginial implementation of StarDist~\cite{Schmidt_2018} for accurate classification while dealing with class imbalance.

\section{Experiments and Results}

We tested two specific hypotheses in our experiments. Firstly, we hypothesized that using the proposed method, pre-training on a related dataset A with class labels derived from the same class hierarchy tree as that of a target dataset B can improve the instance segmentation and classification metrics on the held-out test cases of dataset B compared to training only on dataset B. Secondly, we hypothesized that using the proposed method, domain generalization to a previously unseen dataset C can improve when trained on dataset A and dataset B, as compared to training only on dataset B, where the label sets for the three datasets may be different but are derived from the same class hierarchy tree. For experiments to confirm either hypotheses, we did not discard the last (few) layer(s) after training on dataset A, as is done in transfer learning and multi-task learning. We trained, retained, and re-trained the same last layer by using the proposed adaptive loss functions.

Testing these hypotheses required us to select a test bench, which comprised the following datasets, metrics, DNN architectures, pre-processing methods, training methods and loss functions.

\subsection{Datasets used}

Due to their large size, 40x magnification with clear nuclear details, and a minimal overlap in nuclear classes, we selected three datasets for our experiments -- the Multi-organ Nuclei Segmentation And Classification (MoNuSAC) \cite{9446924} dataset, the PanNuke dataset \cite{gamper2020pannuke}, and the Colorectal Nuclear Segmentation and Phenotypes (CoNSeP) dataset \cite{graham2019hovernet}. More details about these datasets can be found in Section~\ref{sec:datasets}. 

\subsection{Test metric}
Due to its integrated evaluation of instance segmentation and classification, we used panoptic quality (PQ)~\cite{kirillov2019panoptic} to assess our results, which is expressed as follows:
\begin{equation}
\centering
    PQ = \frac{\Sigma_{(p,g) \epsilon TP}IOU(p,g)}{|TP|+0.5|FP|+0.5|FN|}, 
\end{equation}
where $p$ is predicted segment and $g$ is the ground truth segment, $FP$ are false positive predictions, $FN$ are false negative predictions, $TP$ are true positive predictions, and $IoU$ refers to the intersection over union metric. This metric is now widely used for assessing nucleus segmentation and classification.


\subsection{DNN architecture}
We used an instance segmentation and classification architecture used in \cite{Weigert_2022} (which is a modification of the StarDist \cite{Schmidt_2018} model) because it has specific training procedures and post-processing steps for H\&E-stained histology images. It also gives enhanced object localization, leading to higher precision in segmentation, especially of overlapping or closely located nuclei. Additionally, its code repository (made publicly available under BSD license) allowed us to customize the training loss, shape prior, and augmentation techniques.

The architecture consists of a UNet-based backbone network, which can be either a standard UNet \cite{ronneberger2015unet} or other backbones that are similar to it or derived from it.
After the backbone, there are additional convolutional layers which predict a probability map which gives instance segmentation and class probabilities. Additionally, it predicts a distance from the nuclear boundary for multiple directions for each pixel (hence, the name StarDist) to form a polygon.

\subsection{Data preprocessing}

Patches of size 256x256 were extracted from each dataset. Smaller images were appropriately padded. Some patches were overlapping while others were cut-off to fit within 256x256. Images had three channels corresponding to RGB. The ground truth masks had two channels -- the first was the instance segmentation map ranging from 0 to number of nuclei and the second was the classification map ranging from zero to number of classes in the dataset's class label set.


Sometimes due to environmental conditions and staining time, histopathology images face the issue of staining variability of the different dyes such and hematoxylin and eosin that are used to stain the nuclei and the background. This can make it difficult for DNNs to generalize. To combat staining variability, random brightness, hue, and saturation augmentations were performed on the images. To combat class imbalance, geometric augmentations (90 degree flips and rotations) and elastic augmentations were performed more frequently on the less populated classes.

\subsection{Training and Testing details}
We followed the same training approach as described previously \cite{Weigert_2022}. The loss function used was a combination of modified cross entropy (Equation~\ref{eq:SumCE}) and modified focal Tversky loss (Equation~\ref{eq:SumFT}). 
The optimizer used was Adam. We monitored the validation loss for early stopping. Once we finished training the model on one dataset (dataset A) using one instantiation of the modified loss function for a few epochs, we further trained (finetuned) the same model -- without adding or removing any layers or weights -- on the second dataset (dataset B) for a few more epochs by adapting the loss to the second day. The method is flexible enough to take training instances from multiple datasets down to batch-level, but we simplified the procedure to keep the training consistent at an episode (group of epochs) level, where only one dataset was used for training per episode.

\subsection{Results on test subsets}

Table~\ref{tab:TestResults} summarizes the results of testing the first hypothesis that the test results can improve by pre-training on another dataset using the proposed method. Pre-training on another dataset and then fine-tuning for a small number of epochs on our target dataset consistently gave better results for all three target datasets as compared to training only on the target dataset. Thus, our model is able to learn from both datasets even if the labels of the pre-training dataset are different from those of the fine-tuning dataset. Additionally, these results compare favorably with the state-of-the-art for training and testing on various splits of a single dataset~\cite{Schmidt_2018}.

\begin{table}[h]
\caption{Quantitative results on test splits}
\label{tab:TestResults}
\begin{center}
\begin{tabular}{|c|c|c|c|c|c|}
\hline

\textbf{Pre-Train } & \textbf{Epochs}& \textbf{Fine-tune }& \textbf{Epochs}& \textbf{Test }&\textbf{PQ} \\
\hline
CoNSeP& 100& -& 0 &CoNSeP & 0.5404\\
MoNuSAC& 175& CoNSeP& 75 &CoNSeP & \textbf{0.555}\\
PanNuke& 250& CoNSeP& 75 &CoNSeP & \textbf{0.5707}\\

\hline
MoNuSAC& 175& -& 0 &MoNuSAC & 0.5789\\
CoNSeP& 100& MoNuSAC& 130 &MoNuSAC & \textbf{0.5871}\\
PanNuke& 250& MoNuSAC& 130 &MoNuSAC & \textbf{0.6018}\\
\hline
PanNuke& 250& -& 0 &PanNuke & 0.6095\\
CoNSeP& 100&PanNuke & 187 &PanNuke& 0.6056\\
MoNuSAC& 175& PanNuke& 187 &PanNuke& 0.6102\\
\hline
\end{tabular}
\label{tab1}
\end{center}
\end{table}

\begin{figure*}[h]
	\centering
	\begin{subfigure}{0.95\textwidth}
		\includegraphics[width=\textwidth]{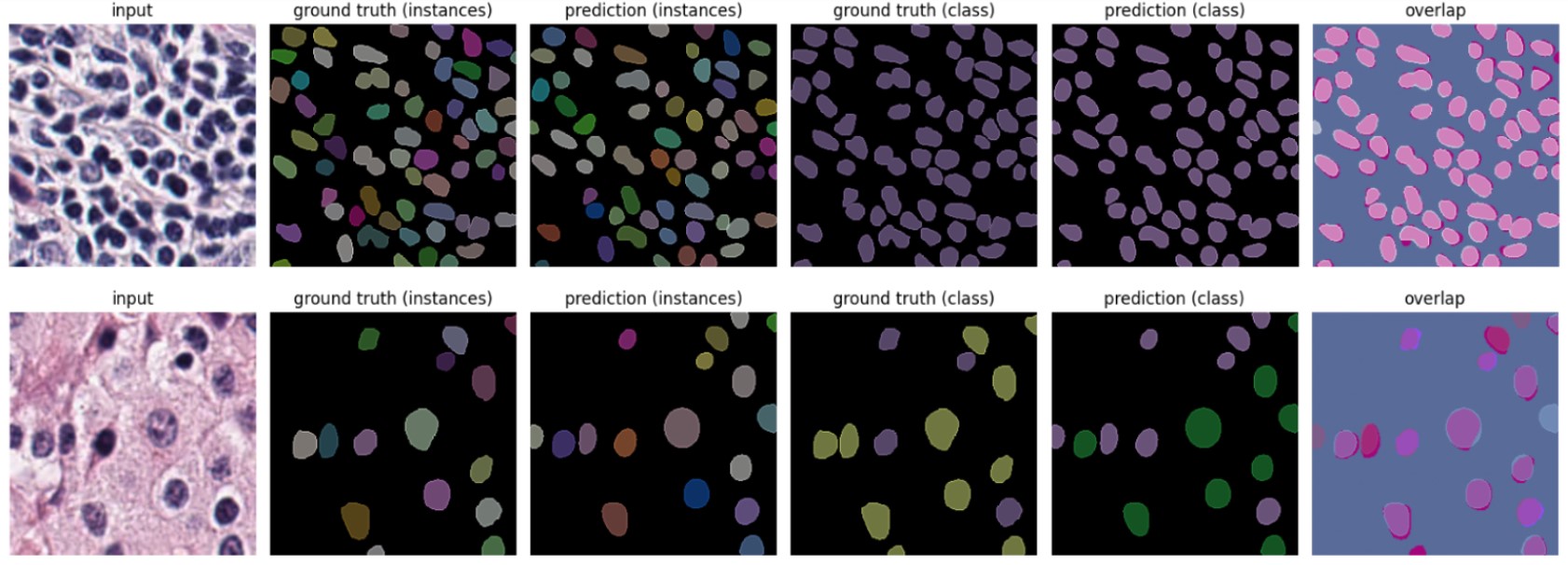}
		\caption{Predictions on MoNuSAC of the model pretrained on PanNuke for 250 epochs followed by fine-tuning on MoNuSAC for 130 epochs}
	\end{subfigure}

	\begin{subfigure}{0.95\textwidth}
		\includegraphics[width=\textwidth]{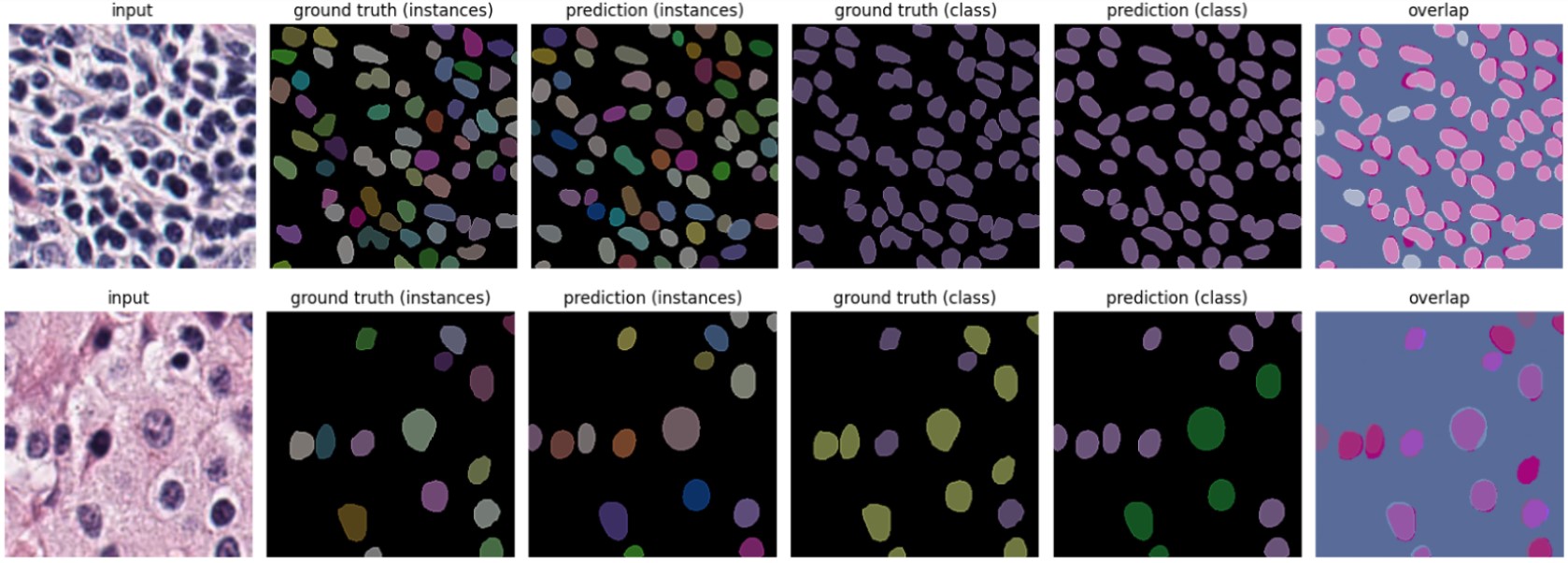}
		\caption{Predictions on MoNuSAC of the model trained on MoNuSAC for 175 epochs before overfitting starts to occur}
	\end{subfigure}

	\caption{A qualitative sample of test split results. }
 \label{fig:qualTestRes1}
\end{figure*}

A sample of qualitative results shown in Figure~\ref{fig:qualTestRes1} also shows better overlap between predicted nuclei and annotations for test images when multiple training datasets are used for training using our method as compared to training on a single dataset.

It is worth noting that the improvement is more pronounced when the pretraining dataset is more generalized and has a super-set of classes and organs as compared to the target dataset. For example, the PanNuke dataset has most of the cell classes present in it. Thus, pre-training on PanNuke and then fine-tuning on other more specialized datasets gives significant improvement in the predictions on those datasets. Pretraining on a smaller specialized dataset like CoNSeP will not benefit the model much, when it is fine-tuned on a broader dataset like PanNuke. Based on this observation and reasoning, the most general dataset in terms of labels can be chosen for pre-training by surveying the classes of the available open source datasets.

\subsection{Evolution of loss upon switching the dataset}

\begin{figure*}[h]
    \centering
\includegraphics[clip,width=15cm,trim=0cm 8cm 7cm 2cm]{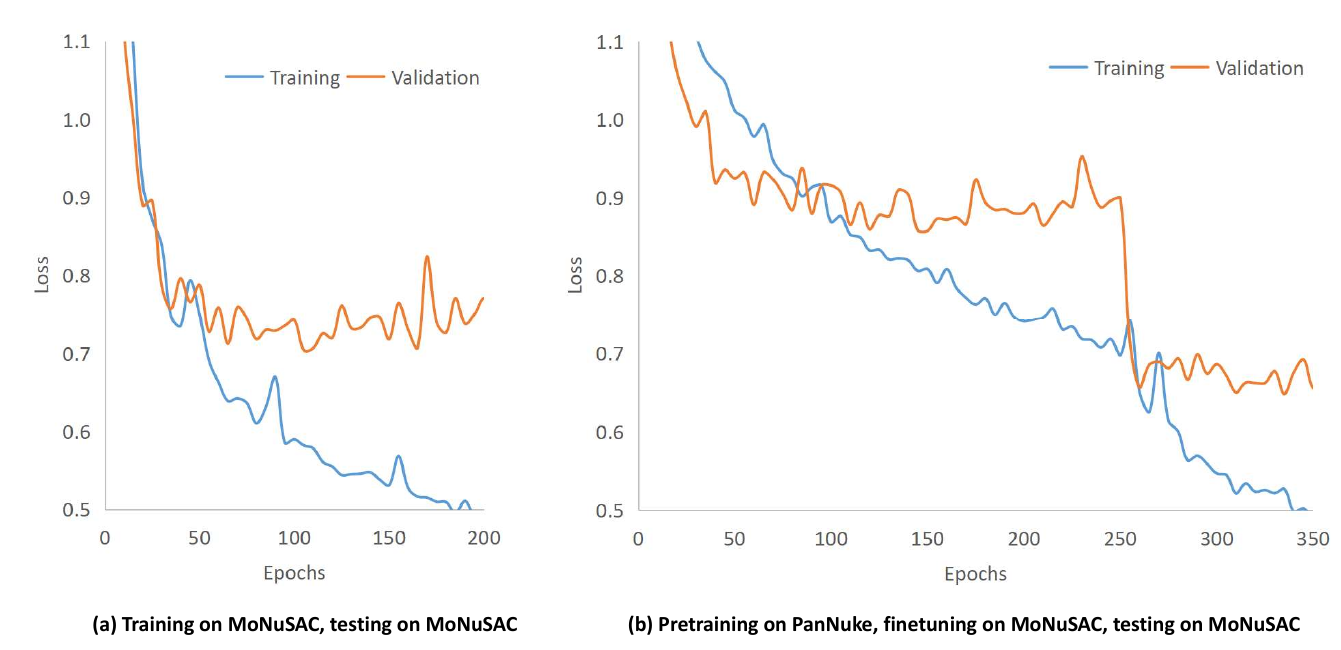}
  \caption{Evolution of training and validation losses for testing on MoNuSAC when (a) trained only on MoNuSAC leading to overfitting, and (b) when pretrained on PanNuke followed by finetuning on MoNuSAC after 250 epochs}\label{fig:LossEvo}
\end{figure*}

Figure~\ref{fig:LossEvo} shows an example of the evolution of the training and validation losses as the training progressed for the MoNuSAC dataset as the target dataset. When trained only on MoNuSAC (case (a)), the model starts to overfit as it can be seen that the validation loss starts to increase. However, when pretrained on PanNuke (case (b)), the validation loss shows a marked further drop when the dataset is switched to the training subset of MoNuSAC as compared to that of case (a). This shows the utility of pre-training using our method.

\subsection{Results on domain generalization}

To test the second hypothesis that domain generalization can improve by training on multiple datasets, we trained the model on the first dataset while monitoring its validation loss to prevent overfitting. After this, we fine-tuned the model on a second dataset. Then we tested on a third dataset, which did not contribute to the training at all. Table~\ref{tab:Generalize} summarizes the results of this experiment. Pre-training on a dataset and then fine-tuning for a small number of epochs on another dataset gives better results on an unseen dataset as compared to training only on the first dataset. Thus, our model is able to consolidate the knowledge of two datasets and show improvement in a domain generalization task.

\begin{table}[h]
\caption{Quantitative results for domain generalization}
\begin{center}
\begin{tabular}{|c|c|c|c|c|c|}
\hline

\textbf{Pre-Train } & \textbf{Epochs}& \textbf{Fine-tune }& \textbf{Epochs}& \textbf{Test }&\textbf{PQ} \\
\hline
CoNSeP& 100& -& 0 &MoNuSAC & 0.4333\\
CoNSeP& 100& PanNuke& 62 &MoNuSAC & \textbf{0.5631}\\
\hline
CoNSeP& 100& -& 0 &PanNuke & 0.4326\\
CoNSeP& 100& MoNuSAC& 43 &PanNuke & \textbf{0.4342}\\
\hline
MoNuSAC& 175& -& 0 &CoNSeP & 0.3444\\
MoNuSAC& 175& PanNuke& 62 &CoNSeP & \textbf{0.4485}\\
\hline
MoNuSAC& 175& -& 0 &PanNuke & 0.3955\\
MoNuSAC& 175&CoNSeP & 25 &PanNuke& \textbf{0.4048}\\

\hline
\end{tabular}
\label{tab:Generalize}
\end{center}
\end{table}

\begin{figure*}[h]

	\centering
	\begin{subfigure}{0.95\textwidth}
		\includegraphics[width=\textwidth]{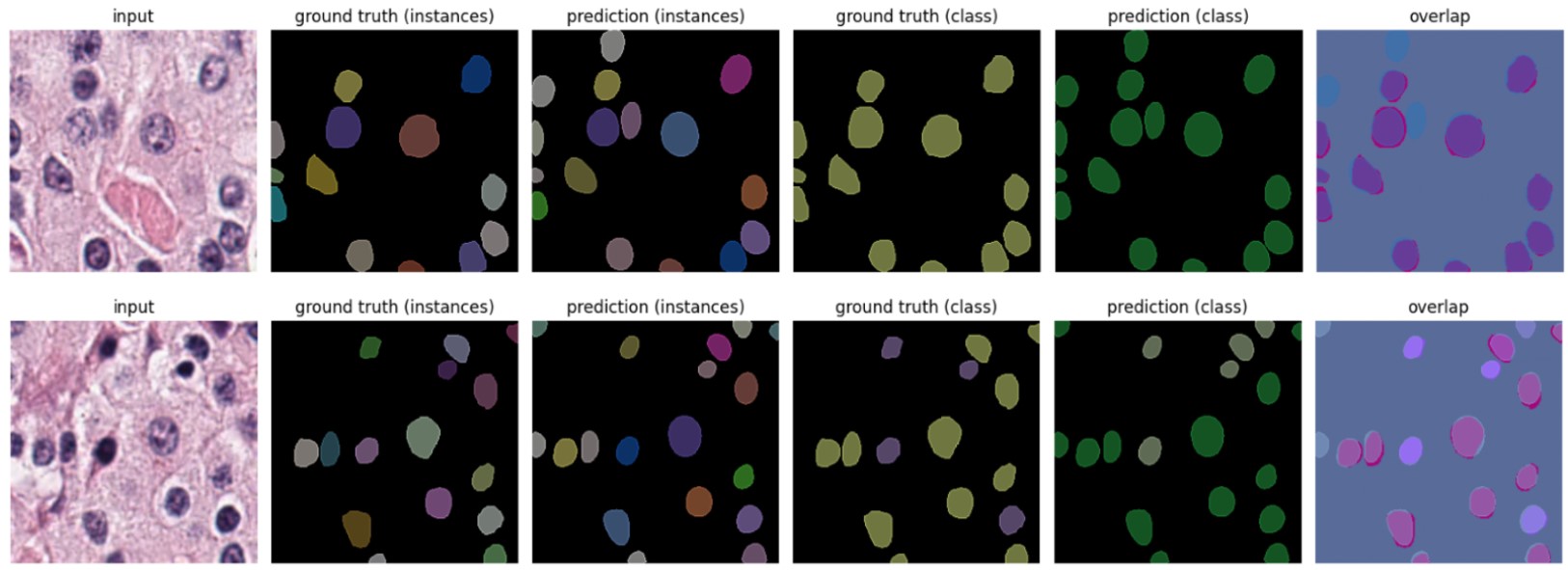}
		\caption{Predictions on MoNuSAC of the model pre-trained on CoNSeP for 100 epochs followed by fine-tuning on PanNuke for 62 epochs}
	\end{subfigure}

	\begin{subfigure}{0.95\textwidth}
		\includegraphics[width=\textwidth]{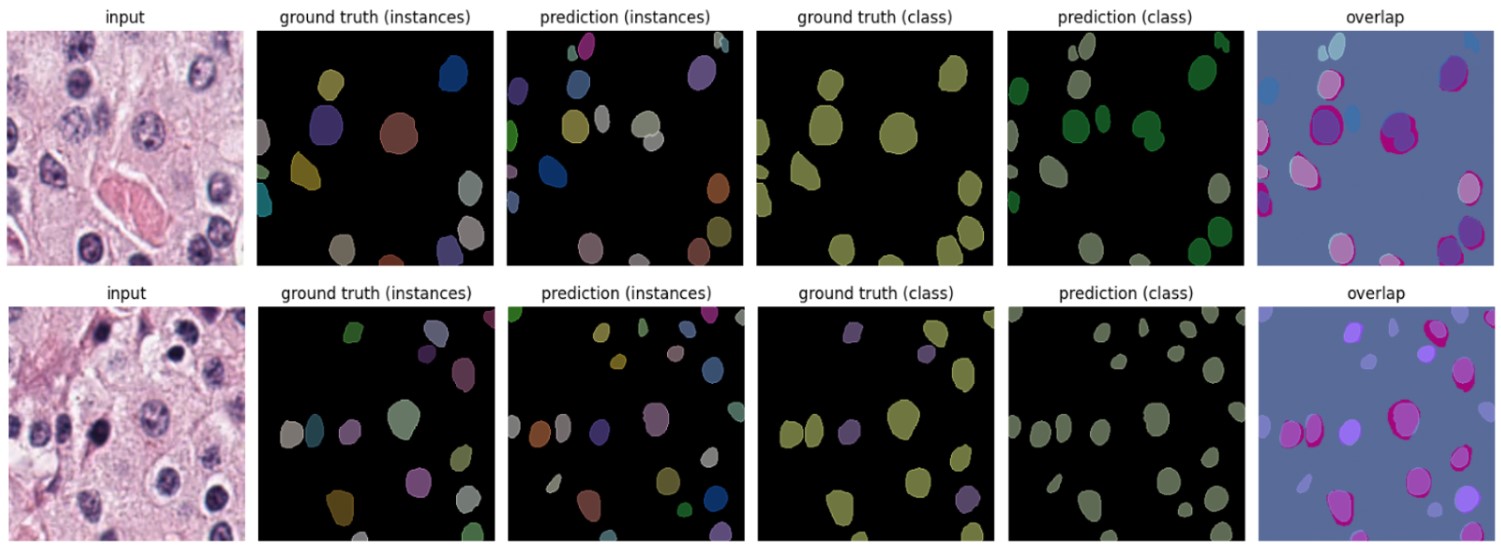}
		\caption{Predictions on MoNuSAC of the model trained on CoNSeP for 100 epochs, before overfitting starts to occur}
	\end{subfigure}

	\caption{A qualitative sample of domain generalization results.}
 \label{fig:qualTestRes2}

\end{figure*}

A sample of qualitative results shown in Figure~\ref{fig:qualTestRes2} also shows better overlap between predicted nuclei and annotations for images from an unseen dataset when multiple training datasets are used for training using our method as compared to training on a single dataset.

We can observe that a more pronounced improvement occurs when the fine-tuning dataset is more generalized and has a super-set of classes and organs as compared to the other datasets. We must take care not to use the most generalized dataset (with a superset of classes) for pretraining because on finetuning with a more specialized dataset, the model loses its accuracy on the unseen dataset instead of benefitting from the fine-tuning. For example, CoNSeP and MoNuSAC are more specialized datasets with classes that have less overlap, but their classes are both subsets of the classes present in PanNuke. In this case, using CoNSeP to finetune the model that was pretrained on PanNuke will lead to decreased performance on MoNuSAC. Now the most general dataset in terms of labels can be chosen by surveying the classes of the available open source datasets.

\section{Conclusion and Discussion}
In this paper, we have proposed a method to combine multiple datasets with different class labels for segmenting and classifying nuclei. We achieved this by creating a hierarchical class label tree to relate the class label sets of different datasets to each other as various cuts of the same tree. We devised a way to combine the losses of the sub-classes, allowing us to train models sequentially on multiple datasets even when the labels are available at a coarser super-class level for some classes and datasets. We demonstrated improved results on test splits and unseen domains (datasets). Our technique can be adapted to other loss functions that involve sum of class probabilities and binary labels, such as focal loss. The principle can also be applied to other application domains (data types), DNN architectures, and tasks such as object detection in settings where different datasets have different label sets which have some overlap with each other. Thus, the method has further scope in various applications and needs to be explored further.






\bibliographystyle{IEEEtran}
\bibliography{name}


\end{document}